\documentclass[sigconf,9pt]{acmart}
\usepackage{tikz}
\usepackage{amsmath}
\usepackage{microtype}
\usepackage{graphicx}
\usepackage{multicol}
\usepackage{xspace}
\usepackage{xcolor}
\usepackage{booktabs} %
\usepackage{adjustbox}
\usepackage{hyperref}
\usepackage{colortbl}
\usepackage{subcaption}
\usepackage{enumitem}
\usepackage{tcolorbox}
\usepackage[linesnumbered,ruled,vlined]{algorithm2e}
\usepackage{balance}
\usepackage[font={small,md},font+=normalfont]{caption} %

\AtBeginDocument{%
  }

\copyrightyear{2024}
\acmYear{2024}
\setcopyright{rightsretained}

\newcommand{\proj}{\emph{Deep Optimizer States}\xspace}

\copyrightyear{2024}
\acmYear{2024}
\setcopyright{rightsretained}
\acmConference[MIDDLEWARE '24]{25th International Middleware Conference}{December 2--6, 2024}{Hong Kong, Hong Kong}
\acmBooktitle{25th International Middleware Conference (MIDDLEWARE '24), December 2--6, 2024, Hong Kong, Hong Kong}\acmDOI{10.1145/3652892.3700781}
\acmISBN{979-8-4007-0623-3/24/12}

\begin{document}

\title[\proj: Towards Scalable Training of Transformer Models Using Interleaved Offloading]{\proj: Towards Scalable Training of\\Transformer Models Using Interleaved Offloading}

\author{Avinash Maurya}
\affiliation{
    \institution{Rochester Institute of Technology}
    \city{Rochester, NY}
    \country{USA}
}
\email{am6429@cs.rit.edu}

\author{Jie Ye}
\affiliation{
    \institution{Illinois Institute of Technology}
    \city{Chicago, IL}
    \country{USA}
}
\email{jye20@hawk.iit.edu}

\author{M. Mustafa Rafique}
\affiliation{
    \institution{Rochester Institute of Technology}
    \city{Rochester, NY}
    \country{USA}
}
\email{mrafique@cs.rit.edu}

\author{Franck Cappello}
\affiliation{
    \institution{Argonne National Laboratory}
    \city{Lemont, IL}
    \country{USA}
}
\email{cappello@anl.gov}

\author{Bogdan Nicolae}
\affiliation{
    \institution{Argonne National Laboratory}
    \city{Lemont, IL}
    \country{USA}
}
\email{bnicolae@anl.gov}
\renewcommand{\shortauthors}{Avinash Maurya et al.}

\begin{abstract}
Transformers and large language models~(LLMs) have seen rapid adoption 
in all domains. Their sizes have exploded to hundreds of billions of parameters and keep increasing.  Under these circumstances, the training of transformers is very expensive and often hits a ``memory wall'', i.e., even when using 3D parallelism (pipeline, tensor, data) and aggregating the memory of many GPUs, it is still
not enough to hold the necessary data structures (model parameters, optimizer
state, gradients, activations) in GPU memory. To compensate, state-of-the-art approaches offload the optimizer state, at least partially, to the host memory and perform hybrid CPU-GPU computations. However, the management of the combined host-GPU memory is often suboptimal and results in poor overlapping between data movements and computations. This leads to missed opportunities to simultaneously leverage the interconnect bandwidth and computational capabilities of CPUs and GPUs. In this paper, we leverage a key observation that the interleaving of the forward, backward and update phases generate fluctuations in the GPU memory utilization, which can be exploited to dynamically move a part of the optimizer state between the host and the GPU
memory at each iteration. To this end, we design and implement \proj, a novel technique to split the LLM into subgroups,
whose update phase is scheduled on either the CPU or the GPU based on our proposed performance model that addresses the trade-off
between data movement cost, acceleration on the GPUs vs the CPUs, and competition for shared resources.
We integrate our approach with DeepSpeed and demonstrate 2.5$\times$ faster iterations over state-of-the-art approaches using extensive experiments.
\end{abstract}

\keywords{Scalable training of large language models, hybrid CPU-GPU I/O performance tuning and middleware, data management for hybrid LLM training, scalable optimization methods for ML}

\ccsdesc[500]{Computing methodologies~Machine learning}
\ccsdesc[500]{Computer systems organization~Heterogeneous (hybrid) systems}
\ccsdesc[500]{Software and its engineering~Data flow architectures}

\maketitle

\section{Introduction}
\label{sec:intro}

Transformers and large language models~(LLMs) have seen increasing 
adoption in various domains ranging from scientific research to
industrial applications~\cite{zhao2023survey}. While traditionally used for creative 
text generation, prompt completion, and comprehension/summarization,
these learning models are successfully tackling multi-modal
data sources, thanks to cross-attention~\cite{MultiModal-Transformers23}.
Additionally, recent initiatives such as LLMs for science (e.g., AuroraGPT~\cite{auroragpt}, ScaleFold~\cite{zhu2024scalefold}, and DeepSpeed4Science~\cite{DeepSpeed4Science23}) are beginning to explore use cases that involve specialized domain-specific languages for tasks, such as, genome sequencing, protein structure prediction, and equilibrium distribution prediction. The versatility and democratization~\cite{rajbhandari2020zero, ren2021zero} of LLMs have led to an unprecedented scale of development across multiple fields.

\paragraph*{\bf Motivation}
In a quest to improve the quality, LLMs are routinely made of billions of parameters with models like GPT-3~\cite{black2022gpt}, LLaMA-2~\cite{touvronLlamaOpenFoundation2023}, and BLOOM~\cite{workshopBLOOM176BParameterOpenAccess2023} requiring hundreds of gigabytes of GPU memory just to store the model parameters. Several predictions anticipate LLMs 
will soon reach trillion scale parameters,
e.g., Google Switch-C (1.6T)~\cite{google-switch}, WuDao 2.0 (1.75T)~\cite{zeng2022glm}, M6-10T~\cite{lin2021m6}, and AuroraGPT~\cite{auroragpt}. 
Despite advances in technologies that enable LLM training to scale (hybrid data-, pipeline- and tensor parallelism, sharding of model parameters and optimizer state, layout and communication optimizations, etc.), the rapid growth in the number of model parameters has resulted in large optimizer states, which has outpaced the available GPU memory, creating a significant ``memory wall'' that makes it challenging to train and run these massive models efficiently~\cite{MemWall23} on limited GPU setups. 
In this paper, we focus on offloading aspects that enable training of moderately complex LLMs (<=20B parameters) on a single node, which is of high value to users that are resource-constrained, e.g., they use a larger HPC system for pre-training but use a fewer number of resource-constrained nodes for quick fine-tuning of LLMs to specialize them for specific tasks~\cite{xia2024understanding}.

\paragraph*{\bf Limitations of State-of-the-Art}
To address the challenge of hitting the memory wall, approaches such as DeepSpeed Offload~\cite{rajbhandari2020zero}, DeepSpeed TwinFlow~\cite{deepspeed-offloadapp}, and Zero-Infinity~\cite{ZeroInfinity-SC21} have explored the idea of moving 
large data structures required during training to the host memory, notably 
the optimizer state. This makes it feasible to train LLMs with a much smaller aggregated GPU memory footprint, albeit at the cost of performance penalty.
Specifically, for commonly used adaptive learning rate optimizers~\cite{you2019large, kingma2014adam} e.g., ADAM, the optimizer state, which includes parameters, momentum, and variance, is stored on the host memory in high FP32 precision, while the forward
pass and backward pass can operate with model parameters in lower FP16
precision to calculate FP16 gradients, which are then flushed to the
host memory and upscaled to FP32 precision. Then, the update of
parameters can proceed directly on the CPU and a downscaled FP16 copy
can be transferred to the GPUs for the next iteration. In this case, 
a critical
bottleneck is \textit{the limited I/O bandwidth between the 
host and GPU memories},
which is constrained by PCIe links (typically in the order 
of 25-50~GB/s). This bottleneck is further exacerbated by 
contention for PCIe links for
inter-node communication needed to implement tensor,
pipeline and data parallelism, which results in additional overhead
during the forward pass (wait for the copy of updated model parameters from the 
host to the GPU memory) and the backward pass (wait to flush the gradients from
the GPU to the host memory). Another important bottleneck is \textit{the
low computational capability of the CPUs}, which are orders of magnitude
slower than the GPUs. For instance, on our testbed (\S~\ref{sec:setup}), the 4$\times$H100 GPUs update $\sim$100~Billion parameters of the model per second~(P/s), while the 192 CPUs update the model at $\sim$8~Billion P/s and copy updated parameters to the GPU at 12 Billion P/s, resulting in 20$\times$ slower updates. Under such circumstances, despite being 
simple and embarrassingly parallel, the operations involved in updating 
the model parameters and the optimizer state lead to a significant
runtime overhead, which otherwise is negligible when running them on 
the GPUs.

\paragraph*{\bf Key Design Ideas and Contributions}
In this paper, we propose \proj to address the two bottlenecks
mentioned above to accelerate the training of LLMs. 
We summarize our contributions as follows:

\begin{enumerate}[topsep=0pt,itemsep=0pt,leftmargin=12pt]

\item We perform a detailed study of 
the behavior of the training iterations
when offloading the optimizer state to the 
host memory. Specifically, we highlight important observations
that drive our proposal: computations remain efficient despite
fine-grain sharding of large optimizer states into subgroups; GPU memory utilization during
the update phase decreases dramatically; and PCIe links are underutilized
during the backward pass and the update phase (\S~\ref{sec:motivation}).

\item We introduce a series of key design principles: 
interleaved offloading of parameter updates on the GPUs; overlapping optimizer subgroup movement and execution across GPU and CPU; efficient placement and movement of gradients for GPU and CPU updates; and PCIe transfers with higher precision to avoid expensive memory allocation for on-the-fly precision conversion (\S~\ref{sec:design}). 

\item We introduce a novel performance model for the update phase to
determine the frequency of GPU offloading
to maximize the overlap with CPU computations and an algorithm to
perform the interleaved CPU-GPU offloading (\S~\ref{sec:design:performance-model}, \S~\ref{sec:design:zoom-algo}).

\item We design and implement \proj, a middleware that integrates our approach into 
widely used LLM training runtimes, namely DeepSpeed~\cite{rasley2020deepspeed} and Megatron~\cite{shoeybi2019megatron}. We insist
on aspects such as the orchestration of background parallelism and interplay
with other existing components (\S~\ref{sec:sys-impl}) for accelerated hybrid CPU-GPU training.

\item We evaluate our implementation in 
a series of extensive experiments in which we train LLMs with up to
20B parameters on resource-constrained setups.
We show significant speed-up in end-to-end LLM training time and up to 3$\times$ faster model
parameter update in a variety of configurations (\S~\ref{sec:evaluation}).
\end{enumerate}

\paragraph*{\bf Limitations of the Proposed Approach}
The proposed approach relies on the model and the optimizer being sharded into smaller subgroups, allowing them to be updated one subgroup at a time. This capability is 
currently implemented in state-of-art LLM training frameworks such as DeepSpeed~\cite{rasley2020deepspeed}, but may not be universally available (e.g., not available in Nanotron~\cite{Nanotron}). Furthermore, this capability
may be allowed only in combination with other capabilities, such as the partitioning
of the subgroups across data-parallel ranks to eliminate redundancy (illustrated
by DeepSpeed ZeRO). In this case, the benefits of dynamic GPU offloading of model updates 
may be offset by the behaviour of complementary capabilities (e.g., redundancy elimination
incurs higher communication overhead compared with replicated data parallelism). However,
most of these limitations are implementation-specific and do not affect the general 
principles.
Furthermore, while \proj accelerates the update phase by leveraging fast GPU-based updates for a fraction of the optimizer states, it is still constrained by slow data movement over PCIe and slow CPU-based updates for remainder subgroups. Therefore, while it mitigates part of the slow CPU-based updates, it does not completely eliminate them to perform as fast as GPU-only updates.

ZeRO-3 Offload, described in \S~\ref{sec:background:zero}, offers additional optimizations for tight memory capacity bound scenarios using quantization, parameter/optimizer offloading to NVMe, activation offloading to NVMe, etc., but 
in this paper,
we specifically focus on and evaluate the scenarios where we have sufficient aggregated GPU memory to store all the model parameters, but not enough to hold the subgroups of optimizer state (consisting of FP32 parameters, momentum, and gradients).

\section{Background and Related work}
\label{sec:background}

\begin{figure*}[t]
    \centering
    \includegraphics[width=0.99\linewidth]{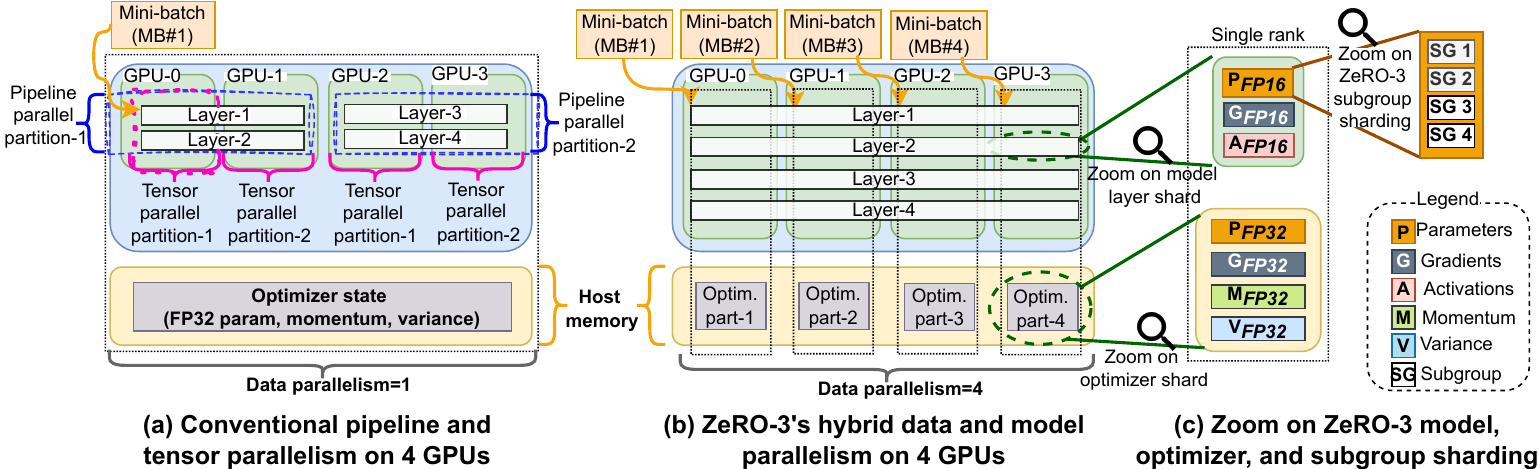}
    \caption{Model parallelism techniques with optimizer state completely offloaded to the host memory: (a) Conventional pipeline and tensor parallelism for a model with 4 layers; (b) DeepSpeed's ZeRO-3 hybrid data and model parallelism; (c) Zoom on the model and CPU-offloaded optimizers of a single data-parallel rank; and subgroup sharding of parameters on each rank. Similar to the sharding of FP16 parameters into 4 subgroups ($SG 1 \dots SG 4$), GPU-resident FP16 gradients and FP16 activations and host-resident FP32 parameters, FP32 gradients, FP32 momentum, and FP32 are sharded in 4 distinct subgroups.}
    \label{fig:hybrid-data-parallelism}
    \Description{An illustration showing model parallelism techniques with optimizer state completely offloaded to the host memory: (a) Conventional pipeline and tensor parallelism for a model with 4 layers; (b) DeepSpeed's ZeRO-3 hybrid data and model parallelism; (c) Zoom on the model and CPU-offloaded optimizers of a single data-parallel rank; and subgroup sharding of parameters on each rank. Similar to the sharding of FP16 parameters into 4 subgroups ($SG 1 \dots SG 4$), GPU-resident FP16 gradients and FP16 activations and host-resident FP32 parameters, FP32 gradients, FP32 momentum, and FP32 are sharded in 4 distinct subgroups.}
\end{figure*}

\paragraph*{\bf Data Parallelism} 
Data parallelism is the most widely used technique to accelerate the training 
of deep learning models~\cite{li2020pytorch, sergeev2018horovod}. It creates replicas of 
the learning model on multiple workers, each of which is placed on a different device and/or
compute node. The input data is randomly shuffled and partitioned among the workers at each epoch. 
During the forward pass, the workers simply process their
mini-batches from the partition of their dataset in an embarrassingly 
parallel fashion. Then, during the backward pass, the model parameters 
are updated based on the average gradients of all replicas (instead of the
local gradients), which effectively synchronizes all replicas to learn
the same patterns from all partitions. Data parallelism leads to accelerated training 
because the partitioning of the input data results in fewer iterations per epoch. 

\paragraph*{\bf Pipeline and Tensor Parallelism}
Pipeline and tensor parallelism are two conventional techniques used to split large models that cannot fit in 
a single GPU memory
by vertically or horizontally sharding the model layers~\cite{jia2022whale, zeng2023distributed, rasley2020deepspeed, guan2019xpipe, fan2021dapple}. As shown in Figure~\ref{fig:hybrid-data-parallelism}(a), tensor parallelism shards individual layers of the model horizontally (denoted by the magenta dotted box), incurring significant communication overheads during the training. On the other hand, pipeline parallelism (denoted by blue dotted boxes in Figure~\ref{fig:hybrid-data-parallelism}(a)), splits the model layers into distinct stages (or pipeline parallel partitions), each of which is placed on a separate GPU. This method requires relatively fewer communications compared to tensor parallelism. Therefore, in real-world training, the tensor-parallelism degree is typically restricted to the maximum number of GPUs available in a single node to leverage high-speed NVLinks, while pipeline stages can be distributed across multiple nodes. Each stage in the pipeline parallel setup can run forward and backward passes of different mini-batches in parallel using gradient accumulation~\cite{smith2022using, huang2019gpipe, gradientAccumPytorch} such that the idle time of GPUs waiting for activations (or gradients) from predecessor (or successor) stages can be minimized by typically using the efficient one-forward one-backward (1F1B) parallelism schedule~\cite{narayanan2019pipedream}. However, when tensor and pipeline parallelism techniques cannot fit the model on GPUs, offloading techniques are used to store the large-sized optimizer state (either fully or partially) 
to the host memory, as shown in Figure~\ref{fig:hybrid-data-parallelism}(a).

\paragraph*{\bf Mixed Precision Training}
\label{sec:background-mixed-precision}
To improve the throughput of training and reduce the GPU memory required for training, LLMs are routinely trained using mixed-precision~\cite{micikevicius2018mixed} 
without negatively impacting the convergence or training accuracy. 
This method allows certain parts of the LLM training to operate in low 16-bit floating point precision e.g., using FP16 or BF16, while others operate in high 32-bit floating point formats e.g., FP32. Real-world LLMs such as BLOOM-176B~\cite{workshopBLOOM176BParameterOpenAccess2023}, OPT-175B~\cite{touvronLlamaOpenFoundation2023}, GPT-3~\cite{black2022gpt}, and GLM-130B~\cite{zeng2022glm} are pre-trained using mixed-precision, wherein the model parameters on the GPU are either in FP16 or BF16 format and the optimizer states are in FP32 format. More specifically, the forward and backward passes can operate with model parameters in FP16 to calculate FP16 gradients, which are then upscaled to FP32 precision and used by the optimizer to compute the updates.

\paragraph*{\bf Hybrid CPU-GPU Optimizer Offloading}
\label{sec:background:hybrid-gpu-cpu}
Several efforts~\cite{chen2018modnn, hildebrand2020autotm, peng2020capuchin, wang2018superneurons} have introduced hybrid training approaches that combine the memory of GPU and other devices for deep learning training. 
For LLMs, to reduce the amount of GPU memory required for training, large optimizer states are either partially or fully offloaded to the host memory or NVMe, using offloading engines, such as ZeRO-Offload~\cite{ren2021zero}, ZeRO-Offload++ or TwinFlow~\cite{deepspeed-offloadapp}, and CoTrain~\cite{li2023cotrain}. When the optimizer state is offloaded to the host memory or NVMe, the low-precision gradients generated on the GPU are moved to the host memory, where it is upscaled to FP32 and consumed by the optimizer for computing updated parameters in high-precision. The updated high-precision parameters are then downscaled and fetched by the GPU to train the next iteration using the updated parameters. Such offloading also accelerates checkpointing (needed at regular intervals for fault tolerance, surviving model spikes, intermediate model analytics, etc.), because the large host-resident optimizer states can be asynchronously flushed 
to persistent storage using several techniques without blocking the GPUs~\cite{datastates-llm,GPUPrefetch-HPDC23,DeepFreeze20,canary-sc22,VELOC-mascots-21}.

To accelerate the update phase for the cases when GPU memory can partially but not fully accommodate the optimizer state, the state-of-the-art LLM training framework, DeepSpeed, offers a partial optimizer offloading optimization using TwinFlow, also known as ZeRO-Offload++. Based on the ``user-defined ratio'', a fraction of the optimizers reside statically on the GPU and the remainder resides on the CPU. Determining the amount of spare GPU memory available for statically storing a subset of the optimizer states is non-trivial and depends on multiple factors, such as model size, parallelism strategy, batch size, individual and aggregated GPU memory capacity, redundancy elimination, and offloading. Due to this complexity, 
the user is typically responsible to profile the pretraining and fine-tune a fixed ratio, as done in TwinFlow~\cite{deepspeed-offloadapp}. However, even with an optimal ratio, the GPU memory
dedicated to storing a part of the optimizer state remains unused during the forward and
backward passes. In our approach, we study the imbalance in the PCIe link and GPU memory utilization throughout the training process and propose a solution to optimize the static hybrid optimizer offloading solutions 
of existing approaches.

\begin{figure*}[t]
\minipage{0.32\textwidth}
    \includegraphics[width=\linewidth]{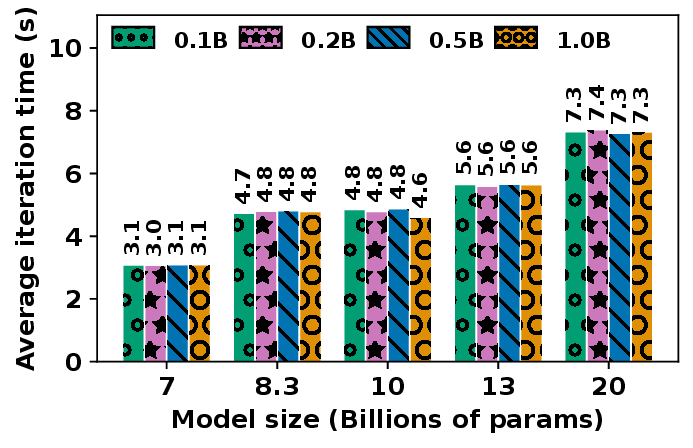}
    \caption{Iteration time for different models when scaling subgroup sizes.}
    \label{fig:scale-subgroup-size}
\endminipage
\hfill
\minipage{0.32\textwidth}
    \includegraphics[width=\linewidth]{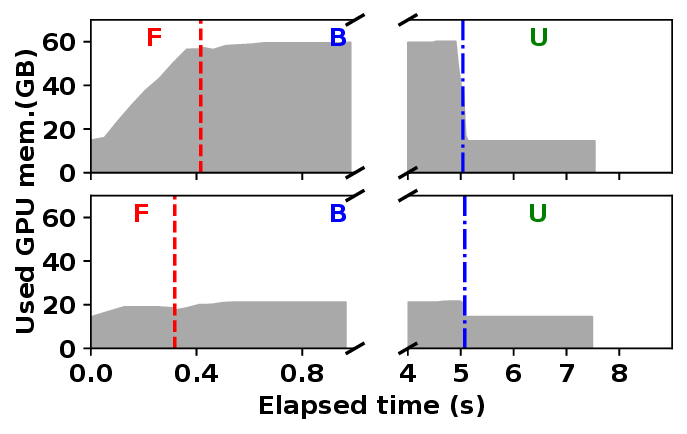}
    \caption{GPU memory util. without (top) and with (bottom) activation checkpoint.}
    \label{fig:mem-util-motivation-20B}
\endminipage
\hfill
\minipage{0.32\textwidth}
    \includegraphics[width=\linewidth]{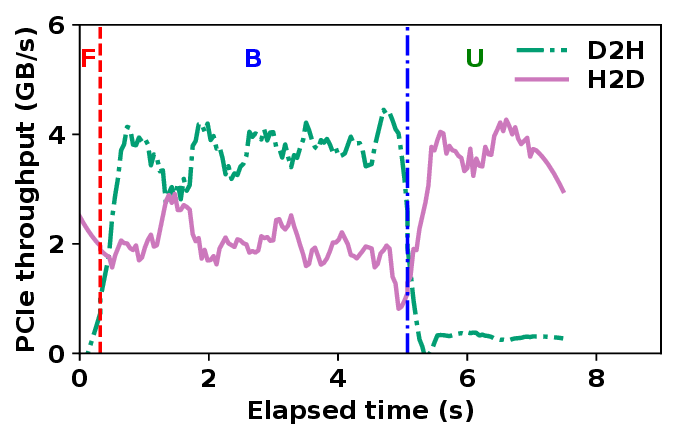}
    \caption{PCIe link util. at different training phases for a 20B parameters model.}
    \label{fig:pci-util-motivation-20B}
\endminipage
\label{fig:motivation-graphs}
\Description{A set of three graphs showing that (a) the iteration duration remains consistent despite the subgroup size; (b) GPU memory utilization fluctuates across training phases because of activations and gradients; (c) PCIe link remains underutilized in all training phases.}
\end{figure*}

\paragraph*{\bf ZeRO: Zero Redundancy Optimizer}
\label{sec:background:zero}
The state-of-the-art LLM training engine DeepSpeed proposes ZeRO~\cite{rajbhandari2020zero, ren2021zero, ZeroInfinity-SC21}, a set of optimizations to eliminate redundancy across data-parallel replicas. To this end, the DeepSpeed runtime supports three different ZeRO stages, namely ZeRO-1, ZeRO-2, and ZeRO-3, each of which incrementally partitions the optimizer state, gradients, and parameters, respectively, across data-parallel replicas. Although ZeRO-1 and ZeRO-2 eliminate optimizer and gradient redundancy across data-parallel ranks, they rely on conventional user-specified pipeline and tensor-parallelism techniques to shard the model across all processes. ZeRO-3 however, adopts a different strategy to eliminate redundancy of the model parameters. As shown in Figure~\ref{fig:hybrid-data-parallelism}(b), it partitions each model layer across data-parallel ranks such that every rank retains only a chunk of a given layer, and performs all-gather of other chunks from the same layer as and when required. This is similar in spirit to tensor-parallelism and incurs significant communication overheads. Nonetheless, a distinguishing feature of ZeRO-3 model sharding is splitting each layer shard further into subgroups, as shown in Figure~\ref{fig:hybrid-data-parallelism}(c). 
Specifically,
if the model consists of $P$ number of parameters and has all the model layers partitioned across $N$ GPUs such that every GPU holds $\sim\lceil P/N \rceil$ parameters, then ZeRO-3 will further divide these $\lceil P/N \rceil$ parameters on every GPU into subgroups of size $SG$ such that every GPU contains~$\sim\lceil \lceil P/N \rceil / SG \rceil$ subgroups. Such subgroup-based sharding allows fine-grained GPU-Host-NVMe movement of model parameters and optimizer states when parameter offloading and optimizer offloading are enabled, respectively. For the ZeRO-3 scenario targetted in this paper, every process owns a unique chunk of the optimizer state and updates it in an embarrassingly parallel fashion. Therefore, no interprocess communication is required in the update phase. In this paper, we exploit such subgroup-style partitioning to efficiently and asynchronously update the optimizer state in chunks using the collective computational throughput of both the CPU and the GPU. For more details about ZeRO-3's design and sharding technique, please refer to ZeRO-Infinity~\cite{ZeroInfinity-SC21}.

\section{Analyzing the Model and System Characteristics during Training}
\label{sec:motivation}

We start by studying the characteristics of the LLM training runtime and the various system resources during training on 4$\times$H100 80~GB GPUs (hardware setup, model description, optimizer, batch size, etc. are detailed in \S~\ref{sec:setup}) using Nvidia Management Library~(NVML)~\cite{nvml}. 
For a more comprehensive analysis please refer to \cite{maurya2024breaking}; below we present the most relevant characteristics for \proj.

\paragraph*{\bf ZeRO-3 Varying Subgroup Sizes}
We first evaluate the impact of varying subgroup sizes on the ZeRO-3 training runtime for different model sizes with the optimizer state completely offloaded to host memory as shown in Figure~\ref{fig:hybrid-data-parallelism}(b). As shown in Figure~\ref{fig:scale-subgroup-size}, we observe that varying the subgroup sizes from 100M to 1B parameters per subgroup does not impact the training iteration for any of the 7B to 20B parameters models. The slight 4\% difference in iteration times can be attributed to uneven partitioning of the model parameters across the GPUs. \emph{Therefore, the subgroup size does not impact the LLM training time.} 

\begin{figure*}[t]
    \centering
    \includegraphics[width=0.99\linewidth]{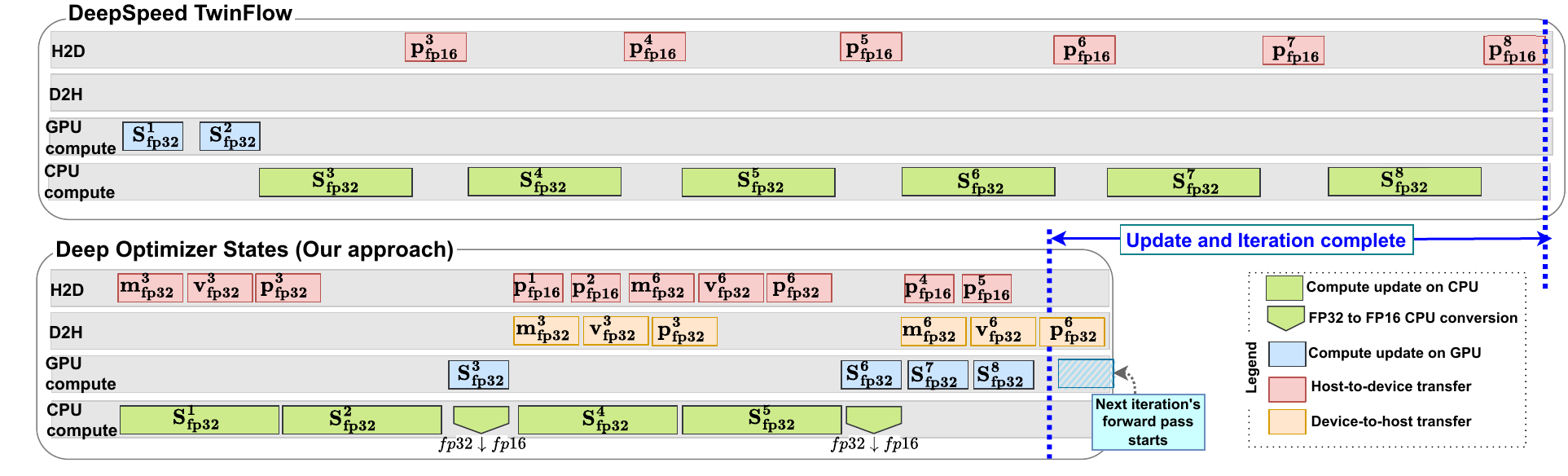}
    \caption{Working of optimizer update step with different approaches for 8 subgroups per GPU (2 subgroups statically residing on GPU). Our approach illustrates an example where 33\% of the updates are scheduled on the GPU.}
    \label{fig:update-working}
    \Description{Illustration showing the working of optimizer update step with different approaches for 8 subgroups per GPU (2 subgroups statically residing on GPU). Our approach illustrates an example where 33\% of the updates are scheduled on the GPU.}
\end{figure*}

\paragraph*{\bf GPU Memory Utilization}
We characterize the GPU memory utilization at different stages, i.e., forward, backward, and update stages of the LLM training. For the 20B parameters model running with optimizer fully offloaded to the host memory, Figure~\ref{fig:mem-util-motivation-20B}~(top) shows the GPU memory utilized for a single GPU when all activations are stored on the GPU during the forward pass. Figure~\ref{fig:mem-util-motivation-20B}~(bottom) shows the memory utilization for the case when activation checkpointing is used to reduce the GPU memory footprint, wherein instead of saving all activations, only a subset of activations at specified intervals (detailed in ZeRO-Infinity~\cite{ZeroInfinity-SC21}~Section-3) are stored on the GPU memory and the remainder are discarded. During the backward pass, the discarded activations are recomputed from checkpoints, resulting in 33\% additional recomputations in the backward pass~\cite{ren2021zero}. When all the activations are stored (Figure~\ref{fig:mem-util-motivation-20B}~(top)), we observe that the GPU memory utilization steeply rises during the forward pass. During the backward pass, these activations are freed and gradients are generated, which get offloaded to host-memory because the optimizer updates are scheduled on the CPU. Lastly, during the update phase, we observe that the GPU only consists of the model parameters, which will get updated once the updates of the CPU offloaded optimizer are complete. A similar trend can be observed for the case when activation checkpointing is used (Figure~\ref{fig:mem-util-motivation-20B}~(bottom)), however, with a lower GPU memory utilization because the activation checkpoints only consume a fraction of the memory in forward pass, which are freed during the backward pass. Irrespective of storing complete activations or activation checkpointing, we observe \emph{significant fluctuations in GPU memory utilization which can be leveraged to store and run a part of the optimizer update step on the GPU}.

\paragraph*{\bf PCIe Link Utilization}
For the 20B parameters model with the optimizer fully offloaded to the host memory, Figure~\ref{fig:pci-util-motivation-20B} shows that both the host-to-device~(H2D) and device-to-host~(D2H) channels are sparsely utilized using <10\% of the peak transfer throughput ($\sim$50~GB/s). During the backward pass, we observe non-negligible H2D and D2H transfers, primarily due to gradient movement. Here, the D2H transfers are caused by the flushing of gradients generated on the GPU by backward pass, which will be used on the host memory to compute model updates by the CPU offloaded optimizer. Surprisingly, during the backward pass, we also observe H2D transfers over the PCIe in Figure~\ref{fig:pci-util-motivation-20B}. This is primarily for faster gradient accumulation; i.e., the gradients are accumulated on the host-memory, and since the accumulation ($old\_grad.add\_(new\_grad)$) operations are magnitudes of order faster on the GPU compared to the CPU, the previously accumulated gradients are 
transferred on the GPU, accumulated on the GPU, and flushed back to the host memory, where the optimizer uses it to compute the updates. Lastly,  during the update phase in Figure~\ref{fig:pci-util-motivation-20B}, we only see H2D transfers, which correspond to fetching the updated parameters from the CPU offloaded optimizer to the GPU for training the next iteration. Therefore, the \emph{PCIe link is underutilized across all the training phases, which can enable partial computation of updates on the GPU.}

\section{System Design}
\label{sec:design}

\subsection{Design Principles}
\label{sec:design:principles}

\paragraph*{\bf Interleaved Optimizer Updates Across GPU and CPU}
The uneven memory consumption and low PCIe link utilization (studied in \S~\ref{sec:motivation}) during different training phases provide an opportunity to exploit the idle GPU memory (released by activations) and PCIe link during the update phase. To exploit this opportunity, during the update phase, parts of the optimizer state can be dynamically fetched on the GPU to compute a fraction of the parameter updates in parallel while the CPU computes updates of the remainder fraction. A key requirement to update the parameters for a given subgroup is to stage its parameters~($p$), momentum~($m$), variance~($v$), and gradients on the target device on which updates are scheduled in FP32 precision (see \S~\ref{sec:background-mixed-precision}). In case the $p$, $m$, $v$, and/or gradients of the subgroup are not present on the target device, the update operation will trigger reads from the slower memory tier (e.g., host memory or NVMe), where the subgroup is offloaded, causing I/O operations in the critical execution path of updates, thereby slowing down the update process. 
By leveraging the fact that adaptive learning rate optimizers, such as Adam~\cite{kingma2014adam}, Adagrad~\cite{duchi2011adaptive}, and RMSProp~\cite{graves2014generating} are embarrassingly parallel, and DeepSpeed ZeRO-3~\cite{rajbhandari2020zero} partitions the optimizer on each process into smaller subgroups (Figure~\ref{fig:hybrid-data-parallelism}(c)), we can perform fine-grained optimizer update scheduling across both GPU and CPU without impacting the consistency of update or introducing computational dependencies (synchronizations) between different subgroups. Furthermore, interleaving does not incur memory allocation and deallocation overheads because on the GPU, memory allocation is handled by PyTorch through lightweight memory pools; and on the host, the memory for all subgroups (except static GPU subgroups) is already pre-allocated and pre-pinned (if enabled) during initialization.

An illustrative example representative of the state-of-the-art hybrid optimizer offloading middleware (e.g., DeepSpeed TwinFlow) is shown in Figure~\ref{fig:update-working}~(top). The optimizer state of a single process is partitioned into 8 subgroups, out of which the first two subgroups ($S^1$ and $S^2$) are statically placed on the GPU, i.e., for the entire training lifetime, the optimizer states corresponding to the two subgroups resides on the GPU memory. Therefore, in Figure~\ref{fig:update-working}~(top), we observe GPU-computations only corresponding to the static GPU-resident subgroups ($S^1$ and $S^2$) at the beginning of the update phase.
The remaining subgroups ($S^3 \dots S^8$) 
are offloaded to the host memory, where the CPU computes the updates (green blocks, last row) and performs H2D transfers (red blocks, top row) of the updated parameters to the GPU naively in blocking fashion. The updated parameters on the GPU are then used in the subsequent training iteration executed on the GPU. 

The interleaved offloading adopted by \proj is illustrated using Figure~\ref{fig:update-working}~(bottom), which schedules 33\% of the subgroups to be updated on the GPU, i.e., for every two subgroups updated on the CPU, one subgroup will be updated on the GPU. The performance model to derive an optimal fraction of optimizer subgroups to be updated by the GPU is described in \S~\ref{sec:design:performance-model}. This interleave-centric design allows for efficient overlap between CPU and GPU computations and asynchronous optimizer-subgroup movement across the PCIe link, which we will detail next.

\begin{figure*}[t]
    \centering
    \includegraphics[width=0.99\linewidth]{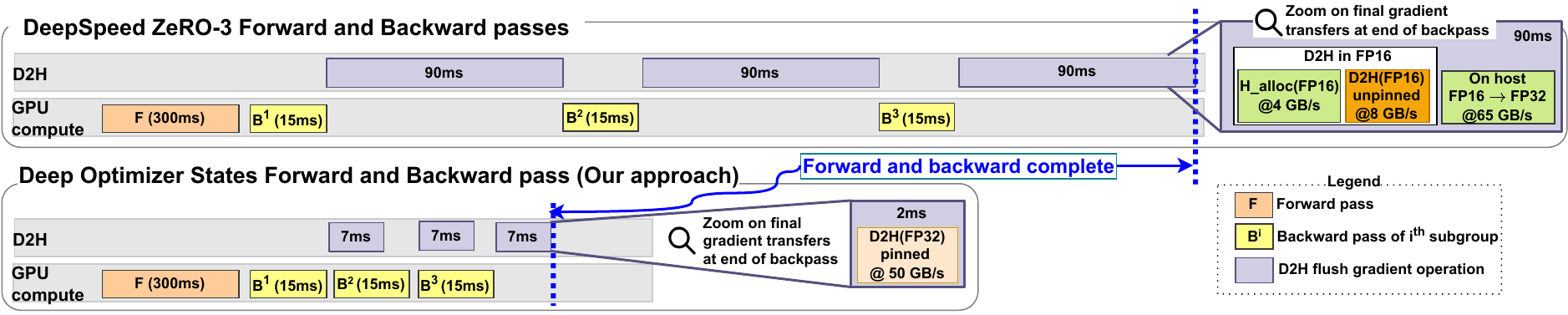}
    \caption{Working of forward and backward passes with different approaches for 3 subgroups per GPU.}
    \label{fig:fwd-bwd-working}
    \Description{An illustration showing the working of forward and backward passes of ZeRO-3 and \proj. ZeRO-3 shows blocking transfer and FP16$\rightarrow$FP32 scaling of the gradients, that we mitigate with \proj.}
\end{figure*}

\paragraph*{\bf Overlapping Optimizer Subgroup Movement and Execution Across CPU and GPU}

The data movement observed when the state-of-the-art middleware enabling hybrid optimizer offload (e.g., TwinFlow~\cite{deepspeed-offloadapp}) runs an update operation is shown in Figure~\ref{fig:update-working}~(top). We observe that after the updates corresponding to a given subgroup $i$ are computed on the CPU, the updated parameters $p^i$ are H2D transferred to the GPU to continue training with updated model parameters in the subsequent iteration. Only when all the subgroups are updated and all the updated parameters transferred to the GPU, the subsequent iteration can begin. Given the embarrassingly parallel nature of optimizer updates~(\S~\ref{sec:background}), the optimizer subgroups can be updated and transferred out-of-order, and do not impact the accuracy of the training. Irrespective of some subgroups statically residing on the GPU, the slow updates using the existing offloading solutions can be attributed to (a) idle CPU when GPU is computing updates of GPU-resident subgroups ($S^1$ and $S^2$); (b) blocking H2D transfer of updated subgroup parameters, i.e., the CPU and GPU remain idle when the parameters corresponding to CPU update subgroup are copied to the GPU; and (c) slow FP32$\rightarrow$ FP16 downscaling of update parameters during H2D transfers (not shown in figure for simplicity).

To mitigate the aforementioned challenges, we propose an overlap-centric design illustrated in Figure~\ref{fig:update-working}~(bottom) for efficient interleaving of CPU and GPU updates. It works as follows: while the CPU computes the update of the initial subgroups ($S^1$ and $S^2$), the optimizer state corresponding to the GPU-scheduled subgroup ($S^3$), including momentum~($m$), variance~($v$), and parameters~($p$), are being prefetched using asynchronous H2D transfers; thereby overlapping CPU computations with GPU subgroup prefetching. Next, the GPU update for subgroup $S^3$ and FP32$\rightarrow$FP16 downscaling of CPU updated parameters ($S^1$ and $S^2$) are done in parallel on the GPU and the CPU, respectively. After this, three operations happen in parallel: (1) H2D transfer of (a) updated parameters of $S^1$ and $S^2$ and (b) prefetching of next subgroup to be updated on the GPU ($S^6$); (2) flushing out (D2H transfer) of the previous subgroup updated on the GPU ($S^3$); and (3) CPU updates of the subsequent subgroups ($S^4$ and $S^5$); thereby exploiting full-duplex D2H and H2D transfers and parallel CPU computations. Furthermore, instead of statically placing the first two subgroups ($S^1$ and $S^2$) on the GPU, we propose to place the last two subgroups ($S^7$ and $S^8$) statically on the GPU to overlap the pending H2D and D2H transfers of previous subgroups updated across host or GPU devices.

As described in \S~\ref{sec:background}, the update phase is executed in an embarrassingly parallel fashion by all processes on a unique chunk of the optimizer which requires no interprocess communications. Consequently, our proposed hybrid CPU-GPU interleaving of subgroup updates using idle PCIe bandwidth does not incur any node-local or cross-node communication overheads. Since the optimizer subgroup movement is process-local and is exclusively dependent on the PCIe throughput, we observe the same speedup during updates at scale, irrespective of the slow cross-GPU interconnect bandwidth.

To enable an efficient overlap of GPU and CPU computations and transfers, we need to devise an optimal fraction of updates to be interleaved and scheduled on the GPU. However, determining this optimal fraction of interleaving is non-trivial and needs to be calibrated based on various factors, such as, the update speed on GPU vs CPU and the PCIe throughput to transfer subgroups back and forth between the GPU and to transfer the updated CPU-based parameters to GPU for the next iteration. Therefore, we complement the interleaved overlap-centric design with a novel performance model (\S~\ref{sec:design:performance-model}) to attain an efficient overlap between update computations and transfers.

\paragraph*{\bf Efficient Management of Gradients for GPU and CPU Scheduled Subgroup Updates}
During the training, the gradients generated during the backward pass on the GPU are used by the optimizer to compute the parameter update. State-of-the-art hybrid optimizer offloading solutions (e.g., TwinFlow), shown in Figure~\ref{fig:update-working}~(top), by default retain the gradients corresponding to the statically GPU-resident subgroups ($S^1$ and $S^2$) on the GPU during the backward pass; and for the remainder of the subgroups, which are scheduled to be updated on the CPU, gradients are offloaded to the host memory during the backward pass. 
In our approach, we extend this design and leverage the GPU memory released by activations (or activation checkpoints) to store the gradients corresponding to the subgroups scheduled for updates on the GPU, which can be known apriori using the lightweight performance model described in \S~\ref{sec:design:performance-model}. In cases where the GPU memory freed by the activations (or activation checkpoints) is not large enough to store the gradients of all GPU-scheduled subgroups, the gradients are offloaded to the host memory and fetched back to the GPU along with the subgroup's optimizer states (FP32 momentum, parameter and variance).

\paragraph*{\bf PCIe Transfers with Higher Precision to Avoid 
Costly
Memory Allocation for On-the-fly Upscaling}
When the models are trained with mixed-precision (\S~\ref{sec:background-mixed-precision}), the gradients generated on the GPU during the backward pass are typically produced in low FP16 precision, whereas the optimizer computes the updates using high FP32 precision gradients. 
Flushing the FP16 gradients from the GPU to the FP32 gradient buffer on the host is non-trivial, which requires both precision conversion (FP16$\rightarrow$FP32) and data movement (D2H transfer). As shown in Figure~\ref{fig:fwd-bwd-working}, for a subgroup size of 0.1B parameters, which generates $\sim$0.2~GB worth of FP16 gradient tensor, the D2H transfer takes place at 2.5~GB/s even when the destination FP32 host gradient buffer is pinned, thereby showing 22$\times$ slower D2H transfer throughput as compared to the peak D2H throughput. When this gradient transfer is zoomed in (rightmost upper block), we observe that this slowdown is because of three different operations involved in the D2H gradient flushing: (1) allocate unpinned memory at $\sim$4~GB/s on the host to hold the FP16 copy of gradients flushed from the GPU; (2) perform the D2H transfer to this unpinned FP16 temporary host buffer at 10~GB/s; and (3) perform FP16 to FP32 conversion on the host at 62~GB/s (as observed in Table~\ref{tab:trf-conv-throughput}). Note that each of the above operations is executed sequentially, thereby stalling the GPU, PCIe, and CPU at different phases throughout the transfer and conversion.

Transferring gradients in FP16 precision in DeepSpeed is adopted to reduce the transfer cost across the PCIe (transfer FP16 instead of FP32). However, even for PCIe Gen 4 interconnect, 
which are widely used in the popular GPUs for training LLMs, e.g., A100,
the achievable D2H throughput is 25~GB/s. This implies that transferring over the PCIe in FP32 would lead to at least 10$\times$ faster gradient flushes as compared to existing 2.5~GB/s throughput. Therefore, in \proj, we adopt to perform chunk-wise in-place on-the-fly conversion from FP16 to FP32 on the GPU (at 1.2~TB/s) and then flush the GPU-resident FP32 gradient chunks to the FP32 pinned gradient host buffer. Table~\ref{tab:trf-conv-throughput} shows the conversion and transfer 
throughputs observed on our testbed used in \S~\ref{sec:setup}.

\begin{table}
    \centering
    \caption{Transfer and conversion throughputs across various devices and data types. G/H represent pinned \underline{G}PU or \underline{H}ost tensors, of 32 ($G_{32}, H_{32}$) and 16 ($G_{16}, H_{16}$) bits, respectively. $\leftrightarrow$ shows the same throughput in both directions.}
    \label{tab:trf-conv-throughput}
    \setlength{\tabcolsep}{2pt}
    \begin{tabular}{|c|c|c|c|c|c|c|}
    \hline
    $G_{32} \leftrightarrow G_{16}$ & $H_{32}\leftrightarrow H_{16}$ & $H_{16} \leftrightarrow G_{16}$ & $H_{32} \rightarrow G_{16}$ & $G_{16} \rightarrow H_{32}$ \\
    \hline
    1.2~TB/s        &   62~GB/s                   & 52~GB/s                   & 8~GB/s                & 4~GB/s         \\
    \hline
    \end{tabular}
\end{table}

\subsection{Performance Model to Determine the Optimal Fraction of Subgroups to be Updated on the GPU}
\label{sec:design:performance-model}
To achieve an efficient overlap of computation and transfers during interleaved optimizer updates, we propose a performance model that computes the ``update stride'', i.e., after how many CPU-based updates should we schedule a subgroup to be updated on the GPU, such that the PCIe link, GPU and CPU are maximally utilized. The key idea of this performance model is to balance the overlap time between CPU-based subgroup updates, GPU-based subgroup updates, and D2H and H2D transfers.

Consider that a single subgroup consists of $S$ number of parameters in high FP32 precision and the CPU to GPU update ratio is $k:1$, i.e., $k$ subgroups are updated on the CPU for every one subgroup updated on the GPU. Furthermore, the update throughput on CPU and GPU are given as $U_c$ and $U_g$ parameters per second, respectively; and the FP32$\rightarrow$FP16 downscaling throughput on the CPU is given as $D_c$ parameters per second. In a given system with H2D and D2H throughputs as $B$ parameters per second, the time to run CPU update and downsampling of $k$ parameters is given by $k*(S/U_c)$ and $k*(S/D_c)$, respectively. For each subgroup updated on the CPU, the downscaled FP16 parameters will be sent to the GPU, resulting in $k*S/(2*B)$ seconds of transfer over the H2D link ($S/2$ instead of $S$ due to FP16 precision). Finally, swapping out the previous optimizer subgroup from the GPU and swapping in the next subgroup on the GPU requires the transfer of FP32 parameters, momentum, and variance, and will require $3*S/B$ seconds of transfer across D2H and H2D PCIe links, respectively.
Specifically, Equation~\ref{eqn:perf_model} formulates the aforementioned computations and data movement to derive the optimal CPU-to-GPU subgroup update ratio. An interesting observation here is that the value of $k$ is not dependent on the subgroup size, therefore, selecting any arbitrary subgroup size results in the same performance improvements of the updates. 
However, smaller subgroups enable the TwinFlow approach to statically store a fraction of optimizer states which is close to the user-supplied ratio, e.g., for a 3B parameters model partitioned in 1B parameters subgroups (i.e., every subgroup is 33\% of the model), if the TwinFlow static GPU-resident optimizer-state ratio is set to 20\%, no subgroup will be scheduled on the GPU; thereby leading to slow updates of all subgroups and GPU underutilization.

\begin{equation} \label{eqn:perf_model}
\begin{split}
k * \Bigl(\frac{S}{U_c} + \frac{S}{D_c}\Bigr) &= \max \left\{
                                                        \begin{aligned}
                                                        &\text{D2H transfers} \\
                                                        &\text{H2D transfers}
                                                        \end{aligned}
                                                    \right\} + \frac{S}{U_g} \\
 &= \max \left\{
                                                        \begin{aligned}
                                                        &\frac{3*S}{B} \\
                                                        &\frac{3*S}{B} + \frac{k*S}{2*B}
                                                        \end{aligned}
                                                    \right\} + \frac{S}{U_g} \\
k &=  \frac{\frac{3}{B} + \frac{1}{U_g}}{\frac{1}{U_c} + \frac{1}{D_c} - \frac{1}{2*B}}
\end{split}
\end{equation}

\subsection{Optimizer Update Scheduling Algorithm}
\label{sec:design:zoom-algo}
Based on the design principles and performance model, the update process of \proj is shown in Algorithm~\ref{algo:opt-update}. In a single update phase, each process invokes the \texttt{run\_update} function using the optimizer subgroups $\langle S \rangle$, the optimal ``GPU update stride'' $k$, i.e., CPUto GPU update ratio derived from the performance model~\S~\ref{sec:design:performance-model}, and the static GPU-resident subgroups $\langle R \rangle$-- configured by the user at runtime, similar to TwinFlow~\cite{deepspeed-offloadapp}.

In Algorithm~\ref{algo:opt-update}, we first check if the given subgroup $i$ is a static GPU resident $\langle R \rangle$ or if it corresponds to the ``update stride'' $k$. Since the subgroups are 0-indexed while the value of $k$ is 1-indexed, we check if $i$ needs to be updated on the GPU through $(i+1)\%k == 0$. While subgroup $i$ is being updated on the GPU, the CPU runs asynchronous downscaling of previous $k-1$ subgroups that were updated on the CPU~(Lines 4-7). If the previous subgroup was updated on the GPU, we launch asynchronous flush-out of the previous subgroup and prefetching of the next subgroup to be updated on the GPU (Lines 8-11). Finally, if the current subgroup was not processed on the GPU, we run the CPU update and enqueue it for future downscaling~(Lines 11-12).

The asynchronous data movement on lines 9-10 of Algorithm~\ref{algo:opt-update} is detailed on lines 13-21. The GPU variables $p\_tmp$, $m\_tmp$, and $v\_tmp$ temporarily store the FP32 parameters $p$, momentum $m$, and variance $v$ for computing a single subgroup's update on the GPU. Since the \texttt{async\_flush\_out} and \texttt{async\_prefetch\_in} operations are launched in parallel to exploit the full-duplex of PCIe, every D2H and H2D is done using a dedicated CUDA stream for transferring the $p$, $m$, and $v$ to establish implicit stream dependency and ensure consistency of flushed-out and prefetched-in subgroups on GPU.

\newcommand\mycommfont[1]{\scriptsize\ttfamily\textcolor{blue}{#1}}
\SetCommentSty{mycommfont}
\begin{algorithm}[t]
    \small
    \caption{Optimizer update scheduling algorithm}
    \label{algo:opt-update}
    \DontPrintSemicolon
    \SetKwInOut{KwIn}{Input}
    \SetKwInOut{KwOut}{Output}
    \KwIn{$\langle S \rangle$: Partitioned optimizer subgroups; $k$: CPU-to-GPU update ratio; $\langle R \rangle$: Static-GPU residents}
    \KwOut{Update target (CPU/GPU) for each subgroup in $S$}
    \SetKwFunction{ScheduleUpdate}{run\_update}
    \SetKwFunction{AsyncFlushOut}{async\_flush\_out}
    \SetKwFunction{AsyncPrefetchIn}{async\_prefetch\_in}
    \SetKwFunction{CPUUpdate}{cpu\_update}
    \SetKwFunction{GPUUpdate}{gpu\_update}
    \SetKwFunction{PrevOnGPU}{prev\_on\_gpu}
    \SetKwFunction{NextOnGPU}{next\_on\_gpu}
    \SetKwFunction{AsyncCPUPrecision}{async\_cpu\_downscale}
    \SetKw{Continue}{continue}
    \SetKw{Or}{or}
    \SetKw{And}{and}
    \SetKwProg{Fn}{Function}{:}{}

    \Fn{\ScheduleUpdate{$\langle S \rangle$, $k$, $\langle R \rangle$}}{
        $fp32\_fp16\_conv = []$ \;
        \For{$i \gets S$} { 
            \uIf{$i \in R$ \Or $(i+1)\%k == 0$  } {
                $\GPUUpdate{i}$ \;
                $\AsyncCPUPrecision(fp32\_fp16\_conv)$ \;
                \Continue \;
                
            } \uElseIf{$i \not\in R$ \And $i\%k == 0$ } { 
                \tcp*{Previous subgroup was updated on GPU}
                $\AsyncFlushOut{\PrevOnGPU{i}}$ \;
                $\AsyncPrefetchIn{\NextOnGPU{i}}$ \;
            }
            $\CPUUpdate{i}$ \; 
            $fp32\_fp16\_conv.append(i)$
        }
    }

    \Fn{\AsyncFlushOut{$x$}}{
        $model_{16}^{G}[x] \gets p\_tmp_{32}^{G}.half()$    \tcp*{D2D: parameter stream} 
        $m_{32}^{H}[x] \gets m\_tmp_{32}^{G}$         \tcp*{D2H: momentum stream}
        $v_{32}^{H}[x] \gets v\_tmp_{32}^{G}$         \tcp*{D2H: variance stream}
        $p_{32}^{H}[x]\gets p\_tmp_{32}^{G}$          \tcp*{D2H: parameter stream}
    }

    \Fn{\AsyncPrefetchIn{$x$}}{
        $m\_tmp_{32}^{G} \gets m_{32}^{H}[x]$ \tcp*{H2D: momentum stream}
        $v\_tmp_{32}^{G} \gets v_{32}^{H}[x]$ \tcp*{H2D: variance stream}
        $p\_tmp_{32}^{G} \gets p_{32}^{H}[x]$ \tcp*{H2D: parameter stream}
    }

\end{algorithm}

\subsection{\proj Implementation}
\label{sec:sys-impl}

We implement \proj as an open-source\footnote{\url{https://github.com/DataStates/artifacts/tree/main/deep-optimizer-states}} middleware for the DeepSpeed ZeRO-3 stage engine. For software packaging, \proj is meticulously engineered and optimized as a Python module that can be enabled and configured through a single JSON entry in the configuration file given to the training runtime. While \proj is designed for Megatron-LM~\cite{shoeybi2019megatron} using DeepSpeed's ZeRO-3 engine (partition model parameters, optimizer, and gradients across data-parallel ranks) approach which uses subgroup-based optimizer sharding, it can be easily extended to other combinations of hybrid parallelization setups, i.e., data-, pipeline-, tensor-parallelism, and ZeRO stages: ZeRO-1 (only partition the optimizer state across data-parallel ranks), ZeRO-2 (partition the optimizer and gradients across data-parallel ranks), by leveraging the embarrassingly parallel runtime of the optimizer updates. We note that our architecture is generic and can be applied with or without DeepSpeed beyond transformer-based language model architectures, e.g., in large vision models, or domain-specific models such as DeepSpeed4Science.
We orchestrate the asynchronous data movement through a modular extension written in C++ and CUDA to enable high-performance transfers and mitigate the limitations of the Python Global Interpreter Lock (GIL). The proposed middleware emphasizes optimizations using dedicated CUDA streams and threads for transfers and asynchronous operations (e.g., downscaling), small pinned buffers for on-the-fly precision conversion which allow for faster DMA transfers, and carefully designed hooks embedded into the training runtime to capture the different phases of training and manage the lifecycle (garbage collection) of tensors across both GPU and the host memory.

\section{Performance Evaluation}
\label{sec:evaluation}

\subsection{Experimental Setup}
\label{sec:setup}

We conduct our experiments on ALCF's JLSE testbed consisting of 4$\times$H100 GPUs with 80~GB HBM3 each (aggregated GPU memory of 320~GB), 2$\times$ Intel Xeon Platinum 8468 processors 
with 
48 CPUs each (total 96 cores, 192 threads), and 2$\times$ Gen4 NVMe of 1.5~TB each. The 512~GB DDR5 RAM is split across 2 NUMA domains, and shared by consecutive GPU IDs, i.e., GPU0 and GPU1 are mapped to NUMA0, and GPU2 and GPU3 are mapped to NUMA1. The GPUs are 
inter-connected
through NVLinks, providing 133~GB/s unidirectional D2D transfer throughput; and every GPU is independently connected to the host with PCIe Gen 5 interface, providing $\sim$55~GB/s unidirectional D2H and H2D throughput for pinned host memory. 
For pageable host memory, the peak unidirectional D2H and H2D throughput are 16~GB/s and 9~GB/s, respectively.

\subsection{Compared Approaches}
\paragraph*{\bf DeepSpeed ZeRO-3} This represents the state-of-the-art technique developed by Microsoft for efficiently training LLMs on GPU-memory-constrained systems. The forward and backward passes of this approach are illustrated in Figure~\ref{fig:fwd-bwd-working}~(top), and the update phase can be illustrated using Figure~\ref{fig:update-working}~(top) with the exception of all subgroups statically residing on the host memory.

\paragraph*{\bf DeepSpeed TwinFlow} This approach is representative of the state-of-the-art hybrid optimizer offloading solution, wherein the optimizer is statically partitioned between host and GPU memory based on the ``user-supplied ratio''. The update phase of this approach is illustrated in Figure~\ref{fig:update-working}~(top).

\paragraph*{\bf \proj} This represents our proposed approach and is highlighted in Figure~\ref{fig:update-working}~(bottom), which uses the design principles and algorithm described in Section~\ref{sec:design}.

\subsection{Methodology and Performance Metrics}

\paragraph*{\bf Models and Datasets}

\begin{table}
    \centering
    \caption{Configuration of models used for evaluations derived from LLaMA2~\cite{touvronLlamaOpenFoundation2023} (7B,13B), Megatron-LM~\cite{shoeybi2019megatron} (8.3B), GPT-10B~\cite{rajbhandari2020zero}, GPT-Neox~\cite{black2022gpt} (20B). The sizes are computed based on ZeRO-Infinity~\cite{ZeroInfinity-SC21}.}
    \label{tab:models}
    \setlength{\tabcolsep}{3pt}
    \begin{tabular}{|c||c|c|c|c|c|c|c|}
    \hline
    Model Size      & 7B    & 8.3B  & 10B & 13B     & 20B   \\
    \hline
    \hline
    Number of layers & 32    & 72    & 50     & 40    & 48    \\
    Hidden dimensions & 4096  & 3072  & 4096   & 5120  & 6144  \\
    Attention heads    & 32    & 24    & 32     & 40    & 64    \\
    FP16 model size (GB) & 24   & 30  & 37    & 46    & 73    \\
    FP32 optimizer (GB)  & 96   & 121 & 150   & 188   & 294   \\
    \hline
    \end{tabular}
\end{table}

The model architectures of the models used in our evaluations, which are based on widely used real-world LLM training, are summarized in Table~\ref{tab:models}. The sizes include the size of FP16 and FP32 gradients model and optimizer states, based on Zero-Infinity~(\S~3)~\cite{ZeroInfinity-SC21}. We restrict our evaluations to 20B parameters models as the next smallest model, LLaMA-33B~\cite{touvronLlamaOpenFoundation2023}, has a larger optimizer state than the DRAM (512~GB) capacity of our testbed.

For our evaluations, we use a subset of the OSCAR-en dataset consisting of 79K records, included in the repository of the Bloom model \cite{workshopBLOOM176BParameterOpenAccess2023}, and use the default LLaMA2 \cite{touvronLlamaOpenFoundation2023} tokenizer for pre-processing the dataset into tokens. Similar to OPT training~\cite{zhang2022opt}, we use the default sequence length of 2048 for all configurations and set the micro-batch size to 1 to avoid OOM errors in any configuration.

\paragraph*{\bf Runtime Configurations}
As described in \S~\ref{sec:background}, ZeRO-3 partitions the model layers across available GPUs in hybrid tensor and data-parallel form. Therefore, we do not use explicit pipeline (unsupported with ZeRO-3) or tensor parallelism. The data-parallel degree is set to 4, which is the maximum number of GPUs in a single node. 
Unless otherwise noted, for all experiments, we use a subgroup size of 100M trainable parameters per subgroup. Although the subgroup sizes do not impact the iteration duration, as observed in Figure~\ref{fig:scale-subgroup-size} or the performance model (\S~\ref{sec:design:performance-model}), as opposed to DeepSpeed's default 1B subgroup size, we choose a smaller subgroup for better static partitioning of optimizer between GPU and CPU with TwinFlow (as detailed in \S~\ref{sec:design:performance-model}). Given the limited GPU memory setup targetted in this paper, similar to Turing-NLG 17.2B, GPT-3 175B, BLOOM-176B~\cite{ZeroInfinity-SC21,workshopBLOOM176BParameterOpenAccess2023}, for all experiments, we used activation checkpointing for reducing the GPU memory utilization at the expense of 33\% additional recomputations during the backward pass. Even if all activations could be saved on the GPU without running OOM, the backward phase would require 33\% less recomputations for all approaches, and we would observe speedup due to overlapping transfers~(Figure~\ref{fig:fwd-bwd-working}). Furthermore, since the activations (or activation checkpointing) are released in the backward phase, they do not impact the update phase and therefore result in the same speedup in the update phase in \proj compared to other approaches.

Throughout our evaluations, we consider that the collective GPU memory is adequate to store the following: (1) FP16 model parameters; (2) activations or activation checkpoints generated by the forward pass; (3) FP16 gradients generated during the backward pass; and (4) at least one FP32 optimizer-state subgroup. Note that a small-sized subgroup consisting of 100M parameters would produce the optimizer state of $3\times sizeof(FP32)\times 100M \approx 1.2~GB$, which can be feasibly obtained by freeing up the activations and FP16 gradients. Based on our performance model (\S~\ref{sec:design:performance-model}), and update and transfer throughputs listed in Table~\ref{tab:trf-conv-throughput} for our testbed~(\S~\ref{sec:setup}), the optimal dynamic ``update stride'' $k=2$, i.e., every alternate subgroup should be updated on the GPU.

\paragraph*{\bf Key Performance Metrics}
We use the following metrics for evaluating the aforementioned approaches: 
(1) the time for computing the forward, backward, and update phases of a single training iteration for various model sizes; (2) update throughput (expressed as billions of parameters updated per second) or the time to update the models of different sizes; and (3) end-to-end training time and TFLOPs achieved by different models. We evaluate these metrics in different scenarios: (a) when the optimizer state is completely offloaded to the CPU memory -- this is representative of scenarios with constrained GPU memory; (b) when the GPU memory is large enough to partially accommodate a fraction of the optimizer state, similar to TwinFlow; (c) when a variable number of CPUs cores are available per GPU to study how the increasing CPU cores impact our proposed approach; and (d) when micro-batch sizes are varied.

\begin{figure*}[t]
\centering
\minipage{0.32\textwidth}
    \centering
    \includegraphics[width=\linewidth]{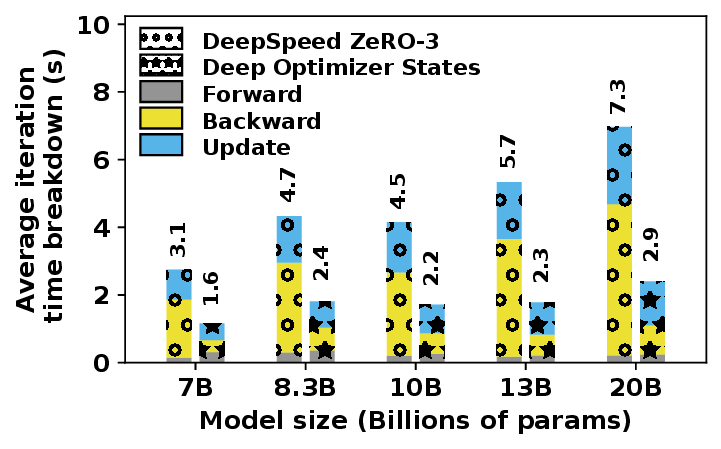}
    \caption{Average iteration time breakdown for different model sizes.}
    \label{fig:diff-models-iter-time}
\endminipage
\hfill
\minipage{0.32\textwidth}
    \centering
    \includegraphics[width=\linewidth]{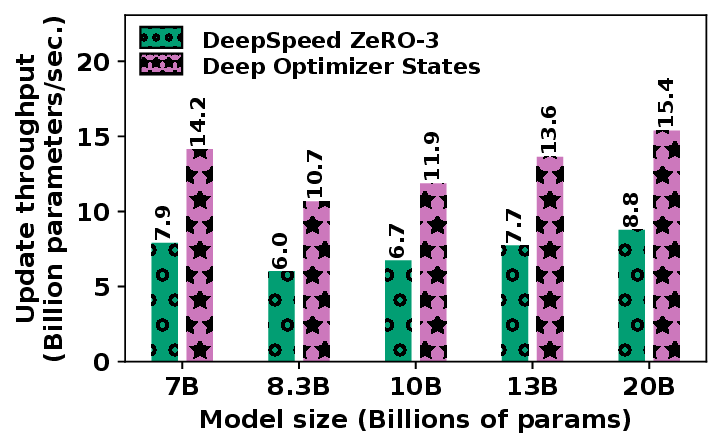}
    \caption{Update throughput for different models.}
    \label{fig:diff-models-update-thru}
\endminipage
\hfill
\minipage{0.32\textwidth}
    \centering
    \includegraphics[width=\linewidth]{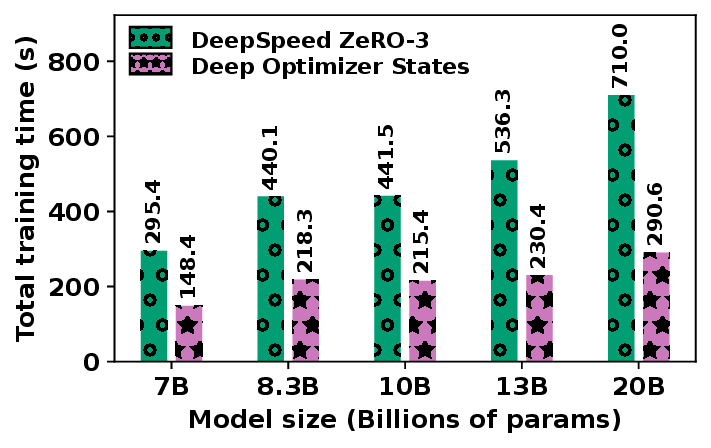}
    \caption{End-to-end runtime for different model sizes.}
    \label{fig:diff-models-end-to-end}
\endminipage
\Description{Three graphs showing the performance improvement of \proj compared to ZeRO-3 for different model sizes, by showing (a) faster average iteration times; (b) update throughput values; and (c) end-to-end runtimes.}
\end{figure*}

\subsection{Experimental Results}

\paragraph*{\bf Optimizer States Completely Offloaded to the CPU Memory}
In our first set of experiments, we evaluate the per iteration time breakdown between forward, backward, and update phases for all the compared approaches when the entire optimizer state resides on the CPU memory for different model sizes listed in Table~\ref{tab:models}. This evaluation studies increasingly large models trained on GPU memory-constrained systems. We run the training for 10 iterations, from which the first 2 iterations are considered warmup and the timings report are the average times observed in 8 iterations. This metric is important because although the subgroups are nearly equally partitioned across all GPU resources, for each subgroup, the backward and update phases invoke blocking allreduce communication collectives (refer~\cite{ZeroInfinity-SC21} for details), because of which the slowest process in the group dictates the iteration time. 

As observed in Figure~\ref{fig:diff-models-iter-time}, the iteration time for larger models with the default DeepSpeed CPU optimizer approach grows linearly in proportion to the model size. The 8.3B parameters model shows higher execution time than the 10B parameters model particularly because of a large number of layers (72) with smaller hidden dimensions (3072) as compared to the other models which consist of at least 4096 hidden dimensions. However, for all model sizes, our proposed \proj shows at least 2$\times$ and up to 2.5$\times$ faster iteration times than DeepSpeed's ZeRO-3 approach. 
When we 
analyze
the speedup obtained with the 20B parameters model with our approach, we observe that asynchronous transfers during the backward pass constitute 1.9$\times$ of the speedup, and the update phase further accelerated the iteration by 60\%, resulting in 2.5$\times$ total speedup.

Next, we analyze the optimizer update step for different model sizes by evaluating the update throughput, which is measured as the total number of optimizer parameters updated per second. As shown in Figure~\ref{fig:diff-models-update-thru}, the update throughput of \proj is 70\% higher than that of ZeRO-3 on average. This is because of efficient overlapping of 50\% the GPU-based updates
using our dynamic offloading middleware. While the update time for increasing model sizes grows proportional to the number of parameters, since the update throughput is a measure of billions of parameters updated per second, Figure~\ref{fig:diff-models-update-thru} shows a near uniform update throughput for different model sizes. 

In the next experiment, we evaluate the end-to-end training time for different model sizes when running for 100 iterations to 
study the impact of asynchronous optimizer movement on subsequent iteration. Specifically, as shown in Figure~\ref{fig:update-working}~(bottom), we study if the overlapping D2H and H2D transfers that spill over the next iteration (marked by a vertical dotted blue line) cause gradual I/O stalls due to limited PCIe and/or host memory read/write throughputs. Figure~\ref{fig:diff-models-end-to-end} shows that the proposed \proj approach achieves nearly the same \textbf{2.5$\times$ speedup in the end-to-end runtime} as observed in per-iteration runtime (Figure~\ref{fig:diff-models-iter-time} for different model sizes), thereby confirming that the overlapping optimizer state movements do not impact the subsequent iterations. 
Another observation we make from Figure~\ref{fig:diff-models-end-to-end} is that running
3$\times$ larger models~(20B) with \proj takes the same time as the 7B parameters model running on state-of-the-art runtimes.

\begin{figure*}[t]
\centering
\minipage{0.32\textwidth}
    \centering
    \includegraphics[width=\linewidth]{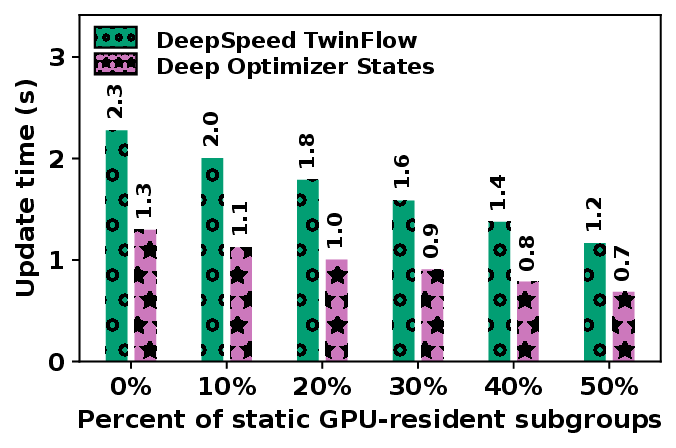}
    \caption{Update time for different TwinFlow ratios for 20B parameters model.}
    \label{fig:diff-twinflow-update-thru}
\endminipage
\hfill
\minipage{0.32\textwidth}
    \centering
    \includegraphics[width=\linewidth]{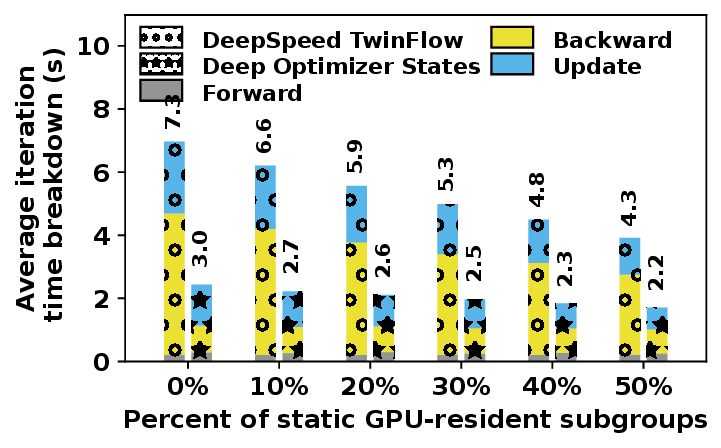}
    \caption{Avg. iteration breakdown for varying TwinFlow ratios, 20B parameters model.}
    \label{fig:diff-twinflow-iter-time}
\endminipage
\hfill
\minipage{0.32\textwidth}
    \centering
    \includegraphics[width=\linewidth]{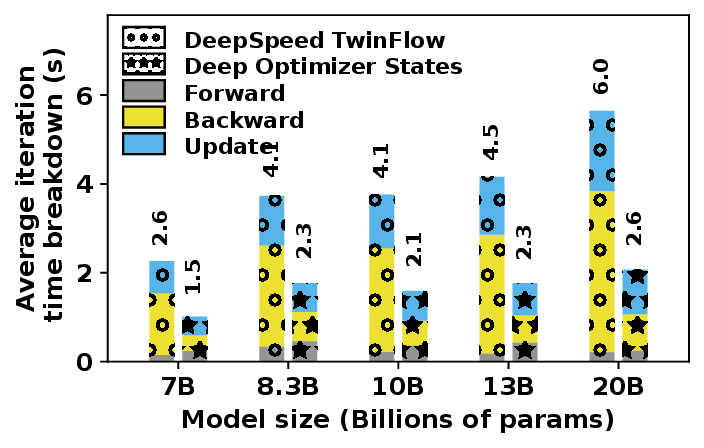}
    \caption{TwinFlow ratio=20\% for different model sizes.}
    \label{fig:diff-model-static-twinflow}
\endminipage
\Description{Three graphs showing the performance improvement of \proj compared to TwinFlow by comparing (a) update time for different TwinFlow ratios for 20B parameters model; (b) iteration breakdown for varying TwinFlow ratios in 20B parameters model; (c) iteration breakdown of different model sizes with static TwinFlow ratio of 20\%.}
\end{figure*}

\paragraph*{\bf Fraction of Optimizer States Statically Resident on the GPU Memory}
In the next series of evaluations, we consider the case when a subset of the optimizer subgroups is statically pinned to the GPU memory. This study shows performance of different approaches when the updates are not completely dependent on slow CPU computations. We use the 20B parameters model as the representative model for subsequent experiments because the longer runtime allows us to better analyze the performance characteristics. 

We evaluate the time for the update phase of the 20B parameters model at varying fractions of optimizer states statically residing on the GPU. Figure~\ref{fig:diff-twinflow-update-thru} shows how the time taken by the update phase decreases with increasing percentage of optimizers statically pinned to the GPU memory. This is because larger proportions of optimizer states residing on the GPU memory lead to faster update computations on the GPU and fewer H2D transfers of the CPU-updated parameters. Irrespective of the proportion of optimizer state on the GPU, we observe at least 1.7$\times$ faster updates with \proj as compared to TwinFlow; thereby showing relevance for efficient training on future GPUs capable of hosting larger proportions of the optimizer state on the GPU memory. 
 
Next, we characterize the performance of a single iteration for the 20B parameters model for varying proportions of optimizer states statically resident on the GPU. Figure~\ref{fig:diff-twinflow-iter-time} shows that our approach achieves $\sim$2$\times$ faster iterations compared to the TwinFlow approach even when the GPU holds as much as 50\% of the optimizer states. An interesting observation from Figure~\ref{fig:diff-twinflow-iter-time} is that \proj performs 40\% faster iterations (3s) at 0\% GPU-offloading as compared to the TwinFlow's 50\% GPU-offloading (4.3s). This means that \textbf{\proj provides 40\% faster iterations at 45\% ($\sim$35~GB per GPU) lower GPU memory utilization compared to the state-of-the-art TwinFlow approach}. 

Lastly, we evaluate increasing model sizes for a fixed TwinFlow static GPU update ratio of 20\%. We select 20\% as a representative GPU-offloading ratio because larger ratios would lead to OOM when running on GPUs with 40~GB HBM, which are typically used in small to medium-scale LLM training, e.g., A100~40~GB GPUs. Compared to the case when the entire optimizer is updated on the CPU with ZeRO-3 (Figure~\ref{fig:diff-models-iter-time}), statically storing 20\% subgroups on the GPU leads to 20\% faster updates as seen in TwinFlow approach in Figure~\ref{fig:diff-model-static-twinflow}. For all model sizes, we observe that our \proj approach outperforms TwinFlow by 1.7-2.3$\times$.

\begin{figure*}[t]
\centering
\minipage{0.32\textwidth}
    \centering
    \includegraphics[width=\linewidth]{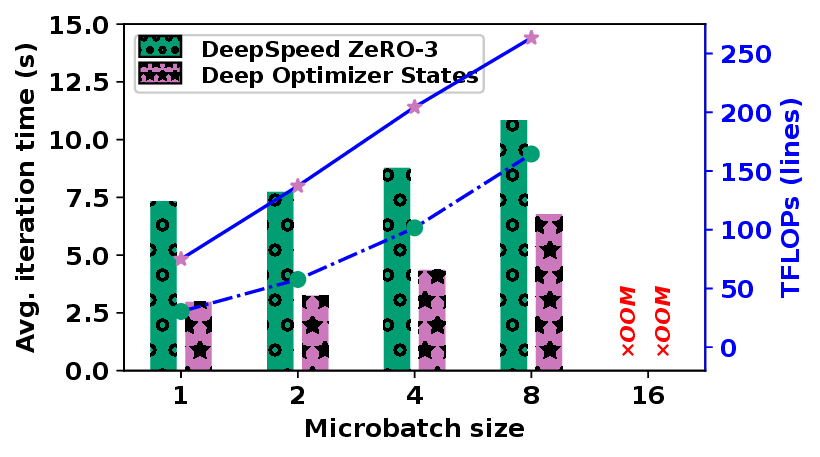}
    \caption{Impact of increasing micro-batch size for the 20B parameters model.}
    \label{fig:diff-mbs}
\endminipage
\hfill
\minipage{0.32\textwidth}
    \centering
    \includegraphics[width=\linewidth]{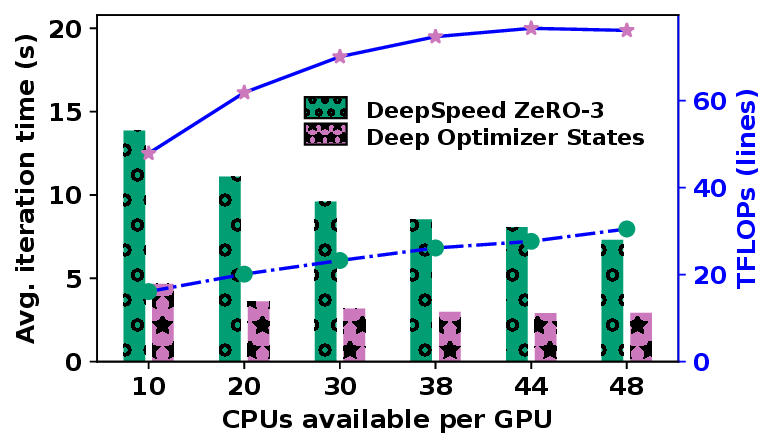}
    \caption{Varying CPU to GPU ratio for the 20B parameters model.}
    \label{fig:diff-cpu-gpu-ratio}
\endminipage
\hfill
\minipage{0.32\textwidth}
    \centering
    \includegraphics[width=\linewidth]{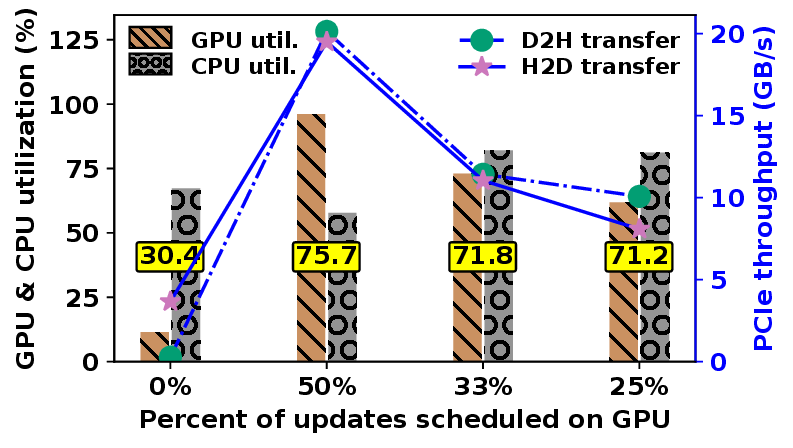}
    \caption{PCIe link and GPU core utilization for 20B parameters model during update.}
    \label{fig:gpu-pcie-util}
\endminipage
\Description{Three graphs showing the performance improvement of \proj over ZeRO-3 for (a) increasing micro-batch size for the 20B parameters model; (b) varying GPU:CPU ratio for the 20B parameters model; and (c) improvement in PCIe link utilization with \proj. }
\end{figure*}

\paragraph*{\bf Increasing Microbatch Sizes}
In the next set of experiments, we evaluate the performance of the 20B parameters model for an increasing microbatch size. To accommodate the largest microbatch on the GPU, the optimizer state resides fully on the host memory during this experiment. We record the total iteration time and the computational throughput achieved (reported as TFLOPs) for an increasing microbatch size per GPU. As shown in Figure~\ref{fig:diff-mbs}, the average iteration time increases linearly in proportion to the growing microbatch size until microbatch of 8 samples, after which, it triggers an OOM error. Although the number of samples increases by 2$\times$ for every x-tick, the iteration time does not grow linearly, leading to higher TFLOPs (reported by the minor y-axis). We observe that the proposed \proj outperforms DeepSpeed's ZeRO-3 by 1.6--2.5$\times$ and scales with increasing microbatch sizes.

\paragraph*{\bf Scaling the CPU Cores per GPU}
We next measure the impact of varying the number of CPU cores available per GPU, which allows us to study different configurations, e.g., ALCF Polaris contains 4$\times$A100 GPUs and 64 CPUs in a single node, AWS p3dn.24xlarge contains 8$\times$ V100 GPUs and 96 vCPUs.
Similar to previous experiments, we consider the optimizer state fully offloaded to the CPU for the 20B parameters model and focus on the performance of a single iteration. 
As observed in Figure~\ref{fig:diff-cpu-gpu-ratio}, for lower CPU to GPU ratio, we observe up to 3$\times$ faster iteration with \proj as compared to ZeRO-3 because fewer CPUs are available to compute the updates on the CPU and GPU-based updating would result in faster updates. As the CPUs to GPU ratio increases, we observe that the average iteration time of both the ZeRO-3 approach and \proj decreases and the achieved TFLOPs increase, and after a point it becomes nearly uniform. This uniform performance with increasing CPU to GPU ratio suggests that once an optimal compute and PCIe overlapping is achieved between the CPU and the GPU, the performance cannot scale any further because of contention (CPU reading/writing updated optimizer states, and concurrent D2H and H2D transfers) on the host memory; thereby making the optimizer update phase dependent on the host memory and PCIe bandwidth.

\paragraph*{\bf CPU, GPU, and PCIe Utilization}
Using the Nvidia Management Library~(NVML)~\cite{nvml}, we perform an ablation study of the compute (GPU, CPU) and transfer (PCIe) resources for the 20B parameters model for varying fractions of updates scheduled on the GPU for a single iteration's update phase. This study highlights how efficiently various node-local resources are utilized to achieve faster updates for offloaded optimizers. Figure~\ref{fig:gpu-pcie-util} shows the GPU and CPU utilization (on the major y-axis) for different fractions of optimizer updates scheduled on the GPU. The case where 0\% of the updates are dynamically scheduled on the GPU is representative of the default DeepSpeed ZeRO-3 approach, which leads to lower CPU utilization ($\sim$70\%) because of blocking H2D transfer of parameters updated on the CPU. Furthermore, the TFLOPs achieved (shown in a yellow box) is $\sim$30 because of negligible GPU and PCIe utilization of 8\% and 2\% respectively. Since the GPU and PCIe utilization metrics were captured using NVML, it reports active GPU utilization even when no kernels are running and only D2D, H2D, or D2H transfers are in progress.
This is because the GPU's copy-engines are actively employed during transfers, and therefore the GPU is not completely idle during transfers. Next, for our proposed approach, when 50\% of the updates are scheduled on the GPU, we observe near-peak GPU utilization of 100\%, and the PCIe links perform D2H and H2D transfers at nearly 40\% of peak unidirectional throughput ($\sim$55~GB/s). However, the CPU utilization for the case of 50\% GPU-scheduled updates goes down to 60\% because of DRAM memory contention between CPU-scheduled updates and concurrent PCIe transfers. 
Nonetheless, even with slightly lower CPU utilization, our approach achieves 75 TFLOPs, which is $\sim$2.5$\times$ faster than the DeepSpeed ZeRO-3 approach. Similarly, when only 33\% and 25\% of the optimizer states are updated on the GPU, our approach significantly outperforms the DeepSpeed ZeRO-3 approach. 
Although for the case of 33\% and 25\% updates scheduled on the GPU, we observe higher CPU utilization compared to the case when 50\% of the updates are scheduled on the GPU, the lower GPU utilization and PCIe transfer results in lower TFLOPS achieved (71 instead of 75), thereby demonstrating that the dynamic optimizer offload problem requires co-optimization of several compute and transfer resources using our proposed performance model (\S~\ref{sec:design:performance-model}).

\begin{figure}[t]
    \centering
    \includegraphics[width=\linewidth]{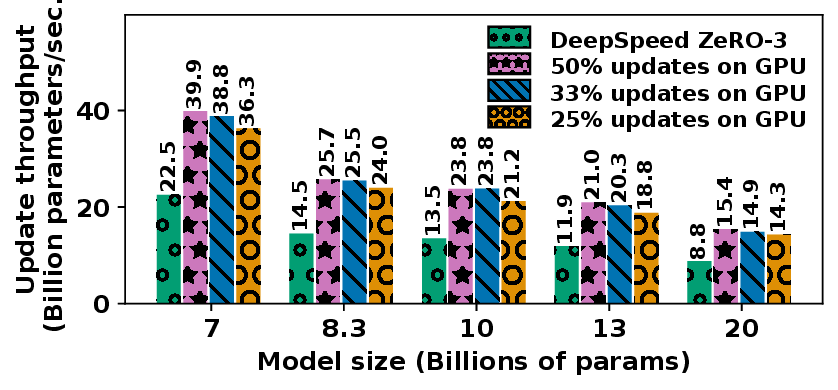}
    \caption{Varying percentages of updates scheduled on the GPU for different model sizes.}
    \label{fig:scale-gaps-diff-models}
    \Description{A graph verifying that our performance model selects the optimal ratio of subgroups to be updated on the GPU.}
\end{figure}

\paragraph*{\bf Verifying the Correctness of Our Performance Model}
We ran several experiments to demonstrate the correctness of our proposed performance model (\S~\ref{sec:design:performance-model}). Figure~\ref{fig:scale-gaps-diff-models} illustrates the update throughput (in billions of parameters updated per second) for different model sizes
and different proportions of updates scheduled on the GPU. For all model sizes, offloading 50\% of the updates on the GPU proves to be an optimal choice. Although we observe near-similar update throughput for the 50\% and 33\% GPU-scheduled updates for the 10B parameters model, the actual update times observed differ by a few seconds which are not significant enough to show a change in the $model\_size$ to $update\_time$ ratio but can lead to significant slowdown when accumulated across thousands of iterations, demonstrating the effectiveness of \proj's performance model. We run similar experiments on a different machine (4$\times$V100 32~GB GPUs, 88 Intel Xeon Gold 6152 cores, and 192~GB host memory) to check that our performance
model is platform-independent. Specifically, for the 7B parameters LLM, we identify the optimal ``update stride''~($k$) using Equation~\ref{eqn:perf_model}. We note the GPU-host transfers~($B$) peaked 3 Billion P/s; GPU~($U_g$) and CPU~($U_c$) update throughputs at 35 Billion P/s and 2 Billion P/s, respectively; and FP32$\rightarrow$FP16 conversion~($D_c$) on host at 8.7 Billion P/s. Using these values in Equation~\ref{eqn:perf_model} results in $k=2$, i.e., every alternate subgroup update should be scheduled on the GPU. Experiments with variable $k$ resulted in update throughputs of 1.67 Billion P/s for $k=3$, 1.62 Billion P/s for $k=4$, and 1.28 Billion P/s for $k=5$, confirming $k=2$ is optimal.

\begin{figure}[t]
    \centering
    \includegraphics[width=\linewidth]{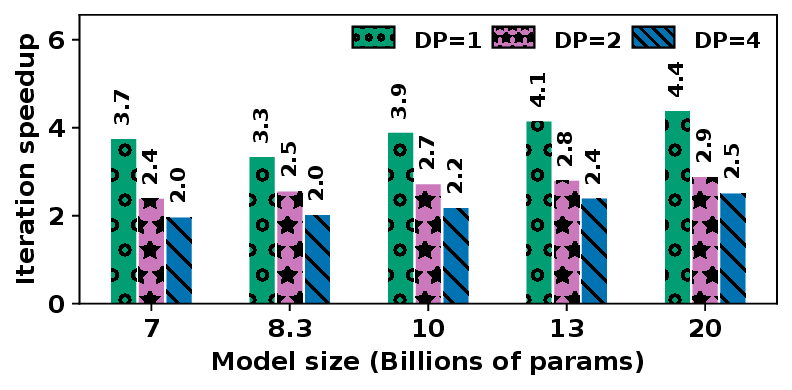}
    \caption{Weak scaling of data-parallelism for different model sizes.}
    \label{fig:scale-dp-diff-models}
\Description{A graph showing the speedup of our approach in a scalability study with increasing data parallelism degrees.}
\end{figure}

\paragraph*{\bf Scaling Data-parallelism Degrees}
In our last set of experiments, we measure the weak scaling performance, i.e., the number of microbatches per GPU remains constant, for an increasing data-parallelism~(DP) degree for different model sizes considering that the entire optimizer state is offloaded to the CPU. Figure~\ref{fig:scale-dp-diff-models} depicts the speedup of \proj as compared to DeepSpeed ZeRO-3 for different model sizes. We observe that for lower data-parallel degrees, \proj obtains up to 4.4$\times$ faster iterations compared to ZeRO-3. As the DP increases, we observe lower speed because (a) increasing DP with weak scaling leads to more training samples processed per iteration; and (b) higher data parallelism leads to model layers sharded across more number of GPUs, which require expensive all-gather operations during the forward and backward passes. 
As a result of both (a) and (b), the longer forward and backward passes diminish the obtained speedup because of faster backward pass and update phases in \proj. Nonetheless, we observe that the iteration speedup is not directly proportional to the degree of data parallelism, and even for higher degrees of data parallelism, \proj shows up to 2.5$\times$ faster iterations demonstrating its efficiency at scale.

\section{Conclusion and Future work}
In this work, we address the problem of slow optimizer updates in LLMs when the large optimizer state is offloaded to the host memory due to limited GPU memory. To mitigate the slow CPU updates for offloaded optimizers, state-of-the-art LLM training frameworks allow partial offloading of optimizer states, resulting in a fraction of the optimizer statically residing on the GPU and the remainder on the CPU. Although this speeds up update performance for the GPU-resident optimizer states, the CPU-based partition slows down the update phase due to limited processing throughput, thereby slowing down the training iteration. To this end, we propose \proj, which leverages the difference in GPU memory and PCIe link utilization during various training phases and performs dynamic scheduling of optimizer updates across both CPU and GPU, resulting in up to 2.5$\times$ faster iterations compared to state-of-the-art solutions.  

Next-generation systems such as Grace Hopper systems feature high-bandwidth (200~GB/s) chip-to-chip~(C2C) interconnect between the CPU and GPU memory, allowing for even faster transfers and computations on the GPU, thereby demonstrating an urgent need for adopting such dynamic interleaved offloading of optimizer states to accelerate training. In the future, we plan to extend and evaluate \proj in multi-node setups for different accelerators and NVMe offloaded optimizer states to speed up the training for even larger models.

\section*{Acknowledgements}
We thank the anonymous reviewers and our shepherd Davide Frey for their constructive feedback to improve this paper. This work is supported in part by the U.S. Department of Energy (DOE), Office of Advanced Scientific Computing Research (ASCR) under contract DEAC02--06CH11357/0F--60169 and the National Science Foundation (NSF) under award no.\  2411386/2411387, 
2106635. Results presented in this paper are obtained using Argonne's ALCF HPC systems, NSF Cloudlab and Chameleon testbed, and Joint Laboratory for System Evaluation (JLSE) at Argonne National Laboratory.

\bibliography{main}
\bibliographystyle{plain}
\balance

\end{document}